\documentclass[arxivdoc]{jmlr}

\usepackage{amsfonts}
\usepackage{float}

\usepackage{longtable}

\usepackage{booktabs}

\usepackage{times,xcolor}
\usepackage{shadethm}

\title{A multi-instance learning algorithm based on a stacked ensemble of lazy learners}

 \author{\Name{Ramasubramanian Sundararajan} \Email{ramasubramanian.sundararajan@ge.com}\\
 \Name{Hima Patel} \Email{hima.patel@ge.com}\\
 \Name{Manisha Srivastava}\thanks{This work was done when this author was employed at GE Global Research} \Email{manisha.s.srivastava@gmail.com}\\  
   \addr GE Global Research\\
John F. Welch Technology Centre, 122 EPIP, Whitefield Road, Bangalore 560066, India}

\begin{document}

\maketitle

\begin{abstract}
This document describes a novel learning algorithm that classifies ``bags'' of instances rather than individual instances. A bag is labeled positive if it contains at least one positive instance (which may or may not be specifically identified), and negative otherwise. This class of problems is known as multi-instance learning problems, and is useful in situations where the class label at an instance level may be unavailable or imprecise or difficult to obtain, or in situations where the problem is naturally posed as one of classifying instance groups. The algorithm described here is an ensemble-based method, wherein the members of the ensemble are lazy learning classifiers learnt using the Citation Nearest Neighbour method. Diversity among the ensemble members is achieved by optimizing their parameters using a multi-objective optimization method, with the objectives being to maximize Class 1 accuracy and minimize false positive rate. The method has been found to be effective on the \textit{Musk1} benchmark dataset.
\end{abstract}

\section{Introduction}
\label{sec:intro}

This document describes a novel learning algorithm that classifies ``bags'' of instances rather than individual instances. A bag is labeled positive if it contains at least one positive instance (which may or may not be specifically identified), and negative otherwise. This class of problems is known as multi-instance learning (MIL) problems.

This setting is applicable in a number of problems where traditional two-class classifiers may face one or more of the following difficulties:

\begin{description}
\item[Precise labeling unavailable] Getting precisely labeled instances is difficult or time-consuming, whereas a precise labeling at a lower level of granularity can be more easily obtained for a larger sample of instances.

Consider the problem of automatically identifying patients who suffer from a certain ailment, based on detection of certain abnormalities in medical images acquired through any appropriate modality (X-ray, CT, MRI, microscopy etc.). Building a model for such automatic identification usually involves creating a labeled sample for training, i.e., having an expert mark out these abnormalities in patients who have said ailment, and creating a training dataset that contains both these images as well as those from normal patients. The model itself is usually some sort of classifier that operates either on images, or regions of interest thereof, or on individual pixels. This method is very well understood and applied in practice.

However, while obtaining ground truth, i.e., labeled examples, certain practical difficulties exist. For instance, the expert may not mark all abnormalities in an image comprehensively and accurately. For instance, if the expert is marking out pink/red coloured rod-shaped objects in a microscopy image of a sputum smear slide to indicate the presence of \textit{Mycobacterium Tuberculosis}, he/she may just mark a few to convey that the patient has the disease, rather than marking every one of them. Also, the marking may be a single pixel inside the object, or an approximate bounding box, rather than a perfect contour of the bacterium. In some cases, the expert may simply mark the image/patient as abnormal rather than mark out the specific region of abnormality, especially in cases where the abnormal region is of diffuse character (e.g. late stage Pneumoconiosis on PA chest x-rays).

These practical issues consequently introduce some label noise in the data, either through unmarked or approximately marked abnormalities. The level of granularity at which the label can be considered reliable is the image/patient itself.

From a traditional classification approach, this throws up two options:

\begin{enumerate}
\item \textit{Learn at a pixel/ROI level in the presence of label noise}. While some classifiers are relatively robust to label noise, their accuracy is generally poorer than when they are learnt without noise. This means that, in cases where the patient has very few regions of abnormality, any error will lead to a misdiagnosis, which is not ideal.

\item \textit{Learn at a patient level}, by characterizing each image (or set of images corresponding to a patient) using a single feature set and then training the classifier using these features and the image-level class labels. The trouble with this approach is that the features themselves are likely to characterize both normal and abnormal regions of the image; depending on the size of the abnormality relative to the size of the image and the nature of the features themselves, the performance of the classifier built using this approach is likely to vary considerably.
\end{enumerate}

A third option, which is proposed in the CAD literature pertaining to multi-instance learning, is to consider each image (or images pertaining to a patient) as an instance bag. The individual instances in the bag may either be pixels or regions of interest (identified using a method with high sensitivity but not necessarily a low false positive rate). These instance bags are labeled using the patient label. In cases where some of the pixels or regions of interest are reliably labeled, we may be able to use this additional information as well.

\item[Classifying sequences] The problem itself is one of classifying a bag, or sequence of instances rather than one of classifying a single instance. Examples include prognostics applications, wherein one may wish to predict whether or not a sequence of events or equipment states is likely to lead to a fault. In this case, although the historical data may have precise labeling as to when the fault occurred, the problem itself is one of classifying a sequence (or “bag”) of machine states. Since the data on which the model is learnt or deployed may not contain time-continuous sequences of states for a particular unit (i.e., data for some periods leading to the fault may be missing), modeling the problem as one of instance bag classification may be a more natural alternative.
\end{description}

The algorithm described in this document is an ensemble-based method, in which the members of the ensemble are lazy learning classifiers learnt using the Citation Nearest Neighbour method. Diversity among the ensemble members is achieved by optimizing their parameters using a multi-objective optimization algorithm, with the objectives being to maximize Class 1 accuracy and minimize false positive rate. 

The organization of this document is as follows: Section \ref{sec:prior} briefly describes the prior work in this area. Section \ref{sec:method} describes the proposed method. Section \ref{sec:apps} describes an application of this method to a benchmark dataset, along with some results. Section \ref{sec:concl} concludes with some directions for further work.

\section{Literature Survey}
\label{sec:prior}

The MIL problem was first discussed in \cite{Dietterich1997}, who also proposed a learning algorithm based on axis-parallel rectangles to solve it. Subsequently, a number of other researchers in the area have proposed MIL algorithms, notably \cite{Wang2000,Zhang2009,Zhou2007,Maron1998,Viola2006,Andrews2002}. For a survey of MIL algorithms, please refer to \cite{Babenko2008,Zhou2004}. In our algorithm, we specifically focus on extending the lazy learning approach proposed in \cite{Wang2000}.

Applications of MIL to real-life problems have been explored in the literature as well. Examples are primarily to be found in the computer-aided diagnostics and image classification area in the form of both papers (\cite{Wu2009,Fung2007,Bi2007}) and patents (\cite{PatentBi2010,PatentKrishnapuram2009,PatentRao2011}). 

The idea of using ensembles for classification in general has been explored extensively in the literature. While \cite{Zhang2009,Zhou2007} discuss the use of ensembles in the MIL context, the method by which the ensemble elements are combined is fairly straightforward (simple voting scheme). \cite{Bi2007} propose the use of a cascaded ensemble of linear SVMs combined using an AND-OR framework, but with the key difference that all elements in the cascade are optimized simultaneously and the execution order of the classifiers is not decided \textit{a priori}.

We approach the ensemble construction problem from the generalized standpoint suggested in \cite{Wolpert1992} -- it is possible that each classifier in the ensemble has learnt a different aspect of the underlying problem; however, combining them may require an additional layer of complexity. We therefore use the stacked generalization approach \cite{Wolpert1992}, wherein a second layer classifier is used to combine the outputs of the first layer ensemble of classifiers. This second layer classifier operates like a single-instance learner, and can therefore be built using any of a variety of standard classifier methods, such as support vector machines, random forests and so on \cite{Breiman1984,Breiman2001,Cristianini2000}.

\section{Methodology}
\label{sec:method}

Consider a set of instance bags $\{B^1, B^2, \ldots B^N\}$, where each bag $B^i$ contains instances $\{X^i_1, \ldots X^i_{N_i}\}$ and is labeled $Y^i \in \{-1,+1\}$ -- we shall refer to them as positive and negative bags. The specific instance-level labels $y^i_j$ for $X^i_j$ may be unknown, except that $Y^i$ is set to $1$ if at least one of $Y^i_j$ is $1$, and $-1$ otherwise. The task of the proposed algorithm is to predict the true label $Y^{new}$ for a an unseen bag $B^{new}$. Let the prediction be denoted by $\hat{Y}^{new}$

The broad steps followed by the proposed classifier are as follows (see figure \ref{fig:MILAlgo}):

\begin{figure}
\includegraphics[width=0.9\textwidth]{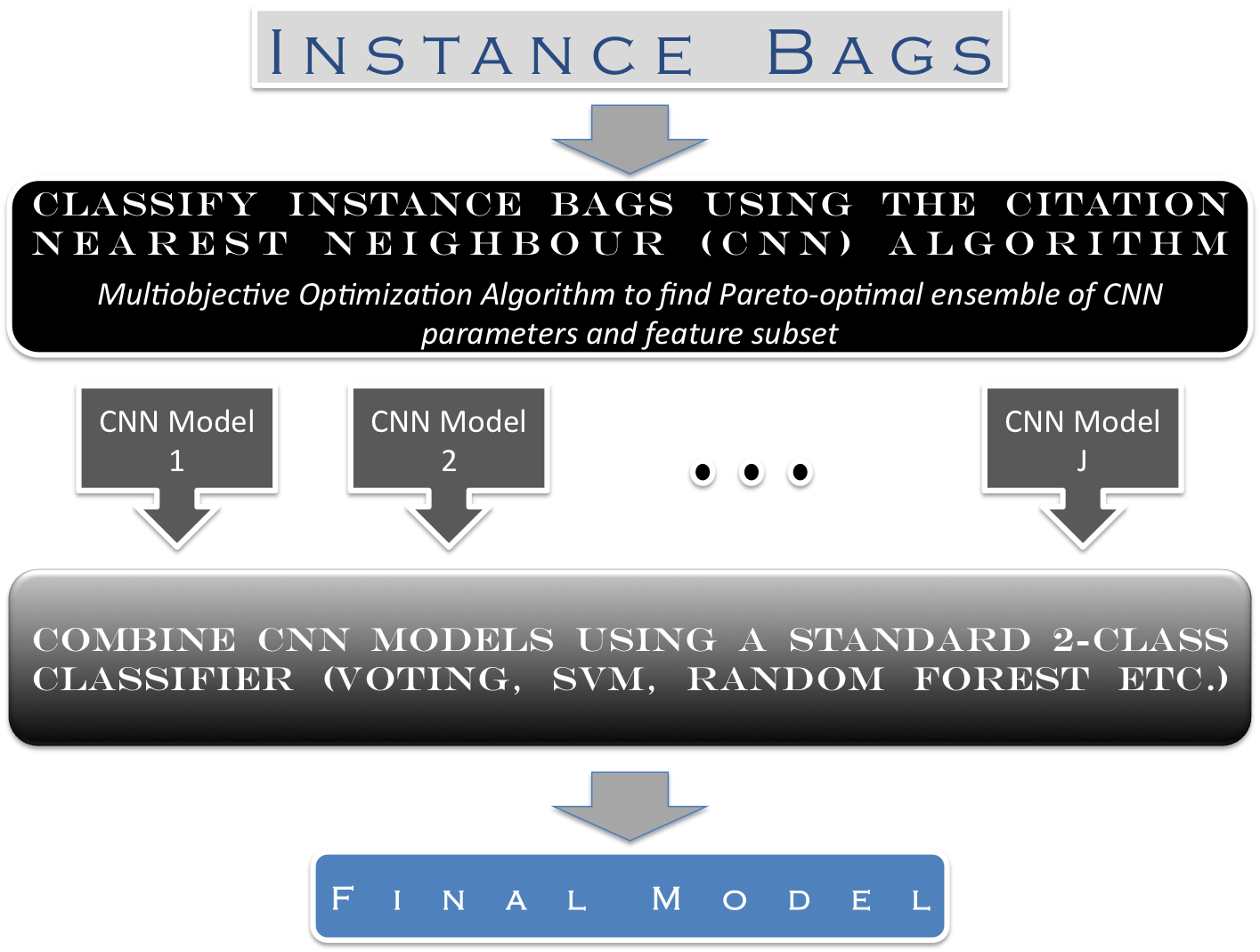}
\caption{Proposed Multi-Instance Learning Classifier}
\label{fig:MILAlgo}
\end{figure}

\begin{enumerate}
\item Use an ensemble of multi-instance classifiers that use the Citation Nearest Neighbour technique. Let the classifiers be denoted by $C_{(j)}, j = 1 \ldots J$, and the predictions from these classifiers for $B^{new}$ be denoted by $\hat{Y}^{new}_{(j)} = C_{(j)}(B^{new})$. Each classifier uses a different set of parameters, so that diversity among the ensemble is maintained. 

\item Combine the predictions $\hat{Y}^{new}_{(1)}, \ldots \hat{Y}^{new}_{(J)}$ using a normal classifier $F$ and return the final prediction, i.e., $\hat{Y}^{new} = F(\hat{Y}^{new}_{(1)}, \ldots \hat{Y}^{new}_{(J)})$.
\end{enumerate}

\begin{algorithm2e}[htbp]
\label{alg:predict}
\caption{Prediction on an unseen instance bag}
\KwIn{\textit{Training sample}: $Train = \{B^1, B^2, \ldots B^N\}$, where each bag $B^i$ contains instances $\{X^i_1, \ldots X^i_{N_i}\}$ and is labeled $Y^i \in \{-1,+1\}$ \\
\textit{CNN classifiers}: $C_{(1)} \ldots C_{(J)}$\\
\textit{Final classifier}: $F$\\
\textit{Unseen instance bag}: $B^{new}$
}
\KwOut{\textit{Prediction for} $B^{new}$: $\hat{Y}^{new}$}

\Begin
{
\begin{enumerate}
\item Apply each CNN classifier $C_{(j)}, j = 1 \ldots J$ to $B^{new}$. Let 

\[
\hat{Y}^{new}_{(j)} \quad = \quad C_{(j)}(Train, B^{new})
\]

be the prediction from the $j^{th}$ CNN classifier.

\item Combine the predictions using the final classifier F. Let 

\[
\hat{Y}^{new} \quad = \quad F(\hat{Y}^{new}_{(1)}, \ldots \hat{Y}^{new}_{(J)})
\]

be the final prediction for $B^{new}$.

\item \textbf{Return} $\hat{Y}^{new}$
\end{enumerate}
}
\end{algorithm2e}

In the following subsections, we describe how this classifier is built. Section \ref{sec:cknn} describes the Citation Nearest Neighbour (CNN) classifier for multi-instance learning. Section \ref{sec:cknnens} describes how an ensemble of CNN classifiers is built, and the predictions of the ensemble combined to get the final prediction. 

\subsection{Citation Nearest Neighbour Algorithm}
\label{sec:cknn}

Our proposed algorithm uses a customized version of a simple, yet effective lazy learning method, namely the Citation Nearest Neighbour (CNN) technique, originally proposed in \cite{Wang2000}.

The CNN technique is simply a nearest neighbour technique with an additional inversion step. We find \textit{references}, i.e., the neighbours of a test bag and note their labels. Similarly, we find \textit{citers}, i.e., those training bags that would consider this test bag a neighbour and note their labels as well. We then compare the number of positive versus negative bags in this calculation and arrive at a final result based on whether the positive bags in references and citers put together outnumber the negative bags. 

The distance between bags is normally calculated using a metric called the minimal Hausdorff distance. Given two bags $A^1 = \{a^1_1 \ldots a^1_p\}$ and $A^2 = \{a^2_1 \ldots a^2_q\}$, the Hausdorff distance is defined as:

\begin{equation}
H(A^1,A^2) = \max(h(A^1,A^2), h(A^2,A^1))
\label{eq:hausdist1}
\end{equation}

where

\begin{equation}
h(A^1,A^2) = \max_{a \in A^1} \min_{b \in A^2} d(a,b)
\label{eq:hausdist2}
\end{equation}

where $dist(a,b)$ is an appropriately defined distance metric between two instances $a$ and $b$.

However, this metric is quite sensitive to the presence of any outlying point on either bag, whose minimum distance from any of the instances in the other bag may be quite high. Therefore, a modified version of Equation (\ref{eq:hausdist2}) is proposed:

\begin{equation}
h_{(d)}(A^1,A^2) = d^{th}_{a \in A^1} \min_{b \in A^2} dist(a,b)
\label{eq:hausdist2a}
\end{equation}

When $d=p$, equations (\ref{eq:hausdist2}) and (\ref{eq:hausdist2a}) are equivalent. When $d=1$, the minimum of individual point distances between the two bags decides the inter-bag distance.

Given a test bag, let $\chi_R$ be the number of \textit{references}, i.e., training sample bags that can be considered as neighbours (i.e., having low Hausdorff distance) to the test bag. This can be found by defining a neighbourhood size $\eta_R$ -- this means that the training bags with the lowest $\eta_R$ distances from the test bag are to be considered as references.

Similarly, let $\chi_C$ be the number of \textit{citers}, i.e., training sample bags that would consider the test bag their neighbour. This can be found by defining a neighbourhood size $\eta_C$ -- this means that the training bags for which the test bag falls within the lowest $\eta_C$ distances from them are to be considered as citers.

Note that the two concepts are not identical -- the test bag may be closest to a particular training bag, but from the standpoint of that training bag, there may be other training bags that are closer to it than the test bag.

Now, let $\chi^+_R$ and $\chi^-_R$ be the number of positively and negatively labeled bags among the references, and $\chi^+_C$ and $\chi^-_C$ be the number of positively and negatively labeled bags among the citers. The predicted label for the test bag $B^{new}$ is $+1$ if:

\begin{equation}
\label{eq:theta}
\frac{\chi^+_R + \chi^+_C}{\chi_R + \chi_C} \geq \theta, \qquad \mathrm{where} \quad \theta = 0.5 \quad \mathrm{typically}
\end{equation}

and $-1$ otherwise. In other words, if there are more positively labeled references and citers for the test bag, its predicted label is positive. See Algorithm \ref{alg:cnn} for the algorithm pseudocode.

\begin{algorithm2e}[htbp]
\label{alg:cnn}
\caption{Citation Nearest Neighbour (CNN) Classifier}
\KwIn{\textit{Training sample}: $Train = \{B^1, B^2, \ldots B^T\}$, where each bag $B^i$ contains instances $\{X^i_1, \ldots X^i_{N_i}\}$ and is labeled $Y^i \in \{-1,+1\}$. Each individual instance $X^i_\ell$ is described by an $m$-dimensional feature vector $\left( X^i_{\ell(1)} \ldots X^i_{\ell(m)}\right)$.\\
\textit{Threshold defining references}: $\eta_R$\\
\textit{Threshold defining citers}: $\eta_C$\\
\textit{Rank used in Hausdorff distance}: $d$\\
\textit{Feature subset}: $S \subseteq \{1 \ldots m\}$\\
\textit{Threshold for classification}: $\theta$\\
\textit{Unseen instance bag}: $B^{tst}$
}
\KwOut{\textit{Prediction for} $B^{tst}$: $\hat{Y}^{tst}$}

\Begin
{
\textbf{function} $\hat{Y}^{tst} = CNN\left(Train, \eta_R, \eta_C, d, S, \theta, B^{tst} \right)$

\begin{enumerate}
\item Calculate the $\Lambda_{(T+1) \times T}$ matrix of pairwise distances between instance bags, where:
	\begin{itemize}	
	\item  \textbf{If} $i \leq T$, \textbf{then} $\Lambda_{i,j} = H(B^i, B^j)$, \textbf{else}  $\Lambda_{i,j} = H(B^i, B^{tst})$
	\item The Hausdorff distance $H(A^1,A^2) = \max(h_{(d)}(A^1,A^2), h_{(d)}(A^2,A^1))$ (see equation \ref{eq:hausdist2a})
	\item The distance metric $dist(a,b)$ between instance pairs from two bags is to be calculated on the feature subset $S$
	\end{itemize}

\item $\chi^+_R = 0$, $\chi^+_C = 0$, $\chi^-_R = 0$, $\chi^-_C = 0$

\item \textbf{For each} $B^i, i = 1 \ldots N$:
	\begin{enumerate}
	\item \textbf{If} $\Lambda_{T+1,i}$ is within the $\eta_R$ smallest values in $\Lambda_{T+1,.}$ and $Y^i = +1$, \textbf{then} $\chi^+_R =  \chi^+_R + 1$
	\item \textbf{If} $\Lambda_{T+1,i}$ is within the $\eta_R$ smallest values in $\Lambda_{T+1,.}$ and $Y^i = -1$, \textbf{then} $\chi^-_R =  \chi^-_R + 1$
	\textbf{If} $\Lambda_{T+1,i}$ is within the $\eta_R$ smallest values in $\Lambda_{.,i}$ and $Y^i = +1$, \textbf{then} $\chi^+_C =  \chi^+_C + 1$
	\item \textbf{If} $\Lambda_{T+1,i}$ is within the $\eta_R$ smallest values in $\Lambda_{.,i}$ and $Y^i = -1$, \textbf{then} $\chi^-_C =  \chi^-_C + 1$
	\end{enumerate}

\item Calculate the classifier score for this example as:

\[
Score \quad = \quad \frac{\chi^+_R + \chi^+_C}{\chi^+_R + \chi^+_C + \chi^-_R + \chi^-_C}
\]

\item \textbf{If} $Score \geq \theta$, \textbf{then} $\hat{Y}^{tst} = +1$ \textbf{else} $\hat{Y}^{tst} = -1$

\item \textbf{Return} $\hat{Y}^{tst}$
\end{enumerate}
}
\end{algorithm2e}

\subsubsection{Customizing the CNN model}
\label{sec:custom}

Optimizing the CNN model typically involves finding the number of references and citers to use (in other words, fix $\eta_R$ and $\eta_C$), as well as the value of the rank $d$ in the Hausdorff distance calculation (empirically, $d=1$ has been found to be effective in most cases). 

However, in our invention we consider the following additional customizations:

\begin{enumerate}
\item In problems where we also know the instance labels (but where it is still beneficial to solve the problem as one of multiple instance learning), we could give higher importance to proximity with a positively labeled instance inside a positively labeled bag.

Since the logic behind the CNN algorithm is that the positive examples across bags are likely to fall close to each other in the feature space, these customizations may allow us to exploit such proximity to a greater extent.

\item While comparing the labels of references and citers put together, we could give higher importance to positive bags than negative ones. This may be useful in situations where the cost of misclassification is asymmetric or where the user is primarily interested in optimizing the accuracy on the positive class, while keeping false positives below an acceptable limit.

\item In situations where the feature set describing each instance is quite large, there is the problem of feature selection in order to arrive at a parsimonious model with good generalization ability. 
\end{enumerate}

Given the above customizations, it is intuitive that one would need to have a process whereby these parameters are appropriately set for the problem in question. Given that our ultimate objective is to create a classifier that can predict  the true label of an unseen test instance bag $B^{new}$ with high accuracy, the ideal combination of parameters ought to be one that maximizes this generalization ability.

In order to estimate generalization ability, we typically use the cross-validation technique. In other words, we take out some of the instance bags in the the training sample (known as the \textit{training subsample}) to use for training, train a model on this part, and test the model on the remaining instance bags (known as the \textit{validation subsample}) as a way of checking its performance on unseen examples. However, in order to avoid sample bias, we need to do this repeatedly and choose different ways to split the training sample into training and validation subsamples. 

A systematic way to do this would be to split the dataset into roughly equal sized chunks (say $k$ chunks). In each iteration, one of the chunks is used as the validation subsample while the other chunks together comprise the training subsample. By doing this with each chunk in turn, we ensure that all examples in the training subsample are used in the validation subsample at some point or another. The average performance of the validation subsamples across these iterations is used as a measure of generalization ability (performance on unseen examples). This approach is referred to as $k$-fold validation in the literature.

An extreme case of this procedure is one where each split contains only one example in the validation subsample -- this method of splitting is repeated as many times as the number of examples in the training sample, and the average performance on the validation subsamples across all these splits is reported as an estimate of generalization ability. This approach is called the leave-one-out strategy (see Algorithm \ref{alg:LOO}). 

Depending on the size of the dataset (i.e., the number of instance bags), one can choose either the $k$-fold validation technique or the leave-one-out strategy described above -- these are the two strategies recommended in our proposed invention.

\begin{algorithm2e}[htbp]
\label{alg:LOO}
\caption{Leave-one-out validation for CNN Classifier}
\KwIn{\textit{Training sample}: $Train = \{B^1, B^2, \ldots B^N\}$, where each bag $B^i$ contains instances $\{X^i_1, \ldots X^i_{N_i}\}$ and is labeled $Y^i \in \{-1,+1\}$. Each individual instance $X^i_\ell$ is described by an $m$-dimensional feature vector $\left( X^i_{\ell(1)} \ldots X^i_{\ell(m)}\right)$.\\
\textit{Threshold defining references}: $\eta_R$\\
\textit{Threshold defining citers}: $\eta_C$\\
\textit{Rank used in Hausdorff distance}: $d$\\
\textit{Feature subset}: $S \subseteq \{1 \ldots m\}$\\
\textit{Threshold for classification}: $\theta$
}
\KwOut{\textit{Class $+1$ (positive) accuracy}: $Acc^+$\\
\textit{Class $-1$ (negative) accuracy}: $Acc^-$\\
\textit{Validation outputs}: $\hat{Y}^{i}, i = 1 \ldots N$
}

\Begin
{
\textbf{function} $\left(Acc^+, Acc^-, \hat{Y}^1, \ldots \hat{Y}^N \right) = LOO\left(Train, \eta_R, \eta_C, d, S, \theta\right)$ 

\begin{enumerate}
\item \textit{Initialize} $n^+ = 0, n^- = 0, a^+ = 0, a^- = 0$
\item \textbf{For} $i \quad = \quad 1 \ldots N$
	\begin{enumerate}
	\item \textit{Set Training subsample}: $TS^i = Train - B^i$
	\item \textit{Set Validation subsample}: $B^i$
	\item \textbf{If} $Y^i = +1$ \textbf{then} $n^+ = n^+ + 1$ \textbf{else} $n^- = n^- + 1$ 
	\item \textit{Call} $\hat{Y}^{i} = CNN\left(TS^i, \eta_R, \eta_C, d, S, \theta, B^{i} \right)$ (see Algorithm \ref{alg:cnn})
	\item \textbf{If} $Y^i = \hat{Y}^{i}$ \textbf{and} $Y^i = +1$ \textbf{then} $a^+ = a^+ + 1$
	\item \textbf{If} $Y^i = \hat{Y}^{i}$ \textbf{and} $Y^i = -1$ \textbf{then} $a^- = a^- + 1$
	\end{enumerate}
\item \textit{Calculate} $Acc^+ = \frac{a^+}{n^+}, \quad Acc^- = \frac{a^-}{n^-}$
\item \textbf{Return} $\left(Acc^+, Acc^-, \hat{Y}^1, \ldots \hat{Y}^N\right)$
\end{enumerate}
}
\end{algorithm2e}

The customization process can therefore be described as follows:

\begin{enumerate}
\item Consider a set of potential combinations of parameters for the CNN model. These include $\eta_R$, $\eta_C$, $d$, relative importance of positive to negative bags while calculating the final label ($\theta$), subset of features to use while calculating the distance metric etc.

\item For each potential combination, estimate the generalization ability using the leave-one-out or similar method.

\item Choose the parameter combination that achieves the best generalization ability.
\end{enumerate}

\subsection{Building the ensemble of CNN classifiers}
\label{sec:cknnens}

When we consider the method described in Section \ref{sec:custom} above to customize a CNN model, two things become obvious:

\begin{itemize}
\item The number of parameters to tune is sufficiently large that the problem of searching through all combinations of parameters may be non-trivial from a computational perspective. Therefore, one may need a smart search algorithm to identify the best combination of parameters from a large possible set.

\item Typically, one wishes to identify a model that maximizes the likelihood of identifying a positively labeled bag correctly, while also minimizing the likelihood of false positives (i.e., negatively labeled bags incorrectly identified). Different parameter combinations are likely to optimize these two metrics in different ways, as they will have different views of the problem space.
\end{itemize}

There is merit in combining diverse views of the same problem to arrive at a more balanced view overall; therefore, we build an ensemble of CNN classifiers. 

We accomplish this by using a multi-objective search heuristic such as NGSA-II \cite{Deb2002} to find the optimal CNN parameters. The search algorithm is asked to find the best set of parameters that optimize the following objectives:

\begin{enumerate}
\item Maximize the likelihood of classifying a positive instance bag correctly
\item Maximize the likelihood of classifying a negative instance bag correctly
\end{enumerate} 

These two objectives are estimated using the leave-one-out method described in Section \ref{sec:custom}.

Note that these two objectives may be in conflict in any problem where perfect separability between the classes is not achievable at the given level of solution complexity. Therefore, the multiobjective search algorithm will throw up a set of candidate solutions, each of which optimizes these two objectives at varying degrees of relative importance. 

Theoretically, the best possible set is known as a \textit{Pareto frontier} of solutions. Any solution in the Pareto frontier cannot be considered superior to another in the frontier (i.e. if it improves on one objective, it loses on another simultaneously), but can be considered superior to all other solutions available. (Note that in this case, when we use the word solution, we refer to a parameter set for the CNN algorithm, and by performance, we refer to the ability to identify positive and negative bags correctly, as measured using the leave-one-out method.)

In practice, a multi-objective optimizer such as NGSA-II will try and arrive at a good approximation to the Pareto frontier, and will output a diverse set of solutions (i.e., parameter combinations for CNN) that optimize the two objectives at varying degrees of relative importance. These solutions constitute the ensemble we wished to construct. The method of combining these solutions is described in Section \ref{sec:stacked}.

\subsubsection{Combining the CNN classifiers in the ensemble}
\label{sec:stacked}

As described earlier, we construct an ensemble of CNN models in order to capture diverse views of the problem to be solved. However, the task lies before us to combine these views. The simplest method of combination would be to let all of these models vote on a test instance bag, and let the majority decide the label. However, it is possible that the optimal method of combination of these diverse views (as represented by the CNN models in the ensemble) calls for a greater degree of complexity than a voting scheme.

Therefore, in our invention, we propose the use of the stacked generalization method \cite{Wolpert1992}, wherein we build a second level classifier $F$, which will combine the predictions of the various CNN models in order to return the final prediction. $F$ can be any two-class classifier such as a support vector machine, random forest etc.

In order to train this classifier, we use the predictions obtained from each member of the ensemble for each instance bag through the validation method described in Section \ref{sec:custom}. One can also choose to optimize the parameters of this classifier using the NSGA-II or similar algorithm, as described in Section \ref{sec:cknnens} above.

\begin{algorithm2e}[htbp]
\label{alg:emoo}
\caption{Optimizing the parameters for CNN using a multi-objective optimization algorithm}
\KwIn{\textit{Training sample}: $Train = \{B^1, B^2, \ldots B^N\}$, where each bag $B^i$ contains instances $\{X^i_1, \ldots X^i_{N_i}\}$ and is labeled $Y^i \in \{-1,+1\}$
}
\KwOut{\textit{CNN classifiers}: $C_{(1)} \ldots C_{(J)}$}

\Begin
{
\textbf{function} $\left(C_{(1)} \ldots C_{(J)}\right) = MOO(Train)$

\begin{description}
\item[Formulation] The problem of finding the optimal ensemble of CNN classifiers is stated as follows:

\begin{equation}
\begin{array}{llll}
\max{Acc^+(C, Train)} & & & \\
\max{Acc^-(C, Train)} & & & \\
\mathrm{where} & & & \\
C & = & \left(\eta_R, \eta_C, d, S, \theta \right) & \mathrm{See Algorithm \ref{alg:cnn}}\\
\left(Acc^+, Acc^-, \hat{Y}^1, \ldots \hat{Y}^N \right) & = & LOO\left(Train, \eta_R, \eta_C, d, S, \theta\right) & \mathrm{See Algorithm \ref{alg:LOO}}
\end{array}
\label{eq:moo-cnn}
\end{equation}

Each candidate solution (CNN classifier) is parameterized in terms of the following variables (see Algorithm \ref{alg:cnn}): $\left(\eta_R, \eta_C, d, S, \theta \right)$. The goodness of each candidate solution is the leave-one-out validation performance of the classifier, in terms of accuracy in identifying positive and negative bags (see Algorithm \ref{alg:LOO}).

\item[Result] A Pareto-optimal set of candidate solutions $C_{(1)} \ldots C_{(J)}$, where every pair $C_{(i)}, C_{(j)}$ of solutions is such that, if $Acc^+_{(i)} > Acc^+_{(j)}$, then  $Acc^-+_{(i)} < Acc^-_{(j)}$, and vice-versa. This means that, without any additional information that allows us to choose between accuracy on positive bags versus accuracy on negative bags, we cannot choose between any of the solutions in the Pareto-optimal set.  
\end{description}
}
\end{algorithm2e}

\begin{algorithm2e}[htbp]
\label{alg:cnn-ens}
\caption{Building a stacked ensemble of CNN classifiers}
\KwIn{\textit{Training sample}: $Train = \{B^1, B^2, \ldots B^N\}$, where each bag $B^i$ contains instances $\{X^i_1, \ldots X^i_{N_i}\}$ and is labeled $Y^i \in \{-1,+1\}$
}
\KwOut{\textit{CNN classifiers}: $C_{(1)} \ldots C_{(J)}$\\
\textit{Final classifier}: $F$
}

\Begin
{
\textbf{function} $\left(C_{(1)} \ldots C_{(J)}, F\right) = StackEns(Train)$

\begin{description}
\item[CNN classifiers] Call a multiobjective optimization algorithm (e.g. NSGA-II) to find an optimal ensemble of CNN classifiers (see Algorithm \ref{alg:emoo}):

\[
\left(C_{(1)} \ldots C_{(J)}\right) = MOO(Train)
\]

\item[Training sample for stacked ensemble] Construct the training sample for generating the second stage classifier, i.e., construct $T2_{N \times J}$ where column $T2_{.j}$ is the set of leave-one-out predictions $\left(\hat{Y}^1_{(j)}, \ldots \hat{Y}^N_{(j)} \right)$ obtained for classifier $C_{(j)}$, generated through the following function call (see Algorithm \ref{alg:LOO}):

\[
\left(Acc^+_{(j)}, Acc^-_{(j)}, \hat{Y}^1_{(j)}, \ldots \hat{Y}^N_{(j)} \right) \quad = \quad LOO\left(Train, \eta_{R(j)}, \eta_{C(j)}, d_{(j)}, S_{(j)}, \theta_{(j)}\right)
\]

Each row $T2_{i.}$ is associated with the class label $Y^i$ for bag $B^i$.

\item[Final classifier] Build a standard $2$-class classifier $F$ using the labeled training set $T2$ generated above.

\item[Return] $\left(C_{(1)} \ldots C_{(J)}, F\right)$
\end{description}
}
\end{algorithm2e}

\section{Empirical validation}
\label{sec:apps}

We demonstrate the utility of our proposed method on the \textit{Musk 1} dataset taken from the UCI Machine Learning repository. This dataset describes a set of $92$ molecules of which $47$ are judged by human experts to be musks and the remaining $45$ molecules are judged to be non-musks. The goal is to learn to predict whether new molecules will be musks or non-musks. However, the $166$ features that describe these molecules depend upon the exact shape, or conformation, of the molecule. Because bonds can rotate, a single molecule can adopt many different shapes. To generate this data set, the low-energy conformations of the molecules were generated and then filtered to remove highly similar conformations. This left 476 conformations. Then, a feature vector was extracted that describes each conformation. 

This many-to-one relationship between feature vectors and molecules lends itself naturally to a multiple instance problem. When learning a classifier for this data, the classifier should classify a molecule as musk if any of its conformations is classified as a musk. A molecule should be classified as non-musk if none of its conformations is classified as a musk \cite{Bache2013}.

\subsection{CNN models}
\label{sec:exp-cnn}

The solution parameters to be optimized are the CNN model parameters, as well as the feature subset used to compute distance between instances. Since this is a large-scale multi-objective optimization problem with objectives where the gradient is ill-defined, we use a direct search method such as a multi-objective genetic algorithm to solve it. Specifically, we use the Non-dominated Sorting Genetic Algorithm II (NSGA-II) to optimize the CKNN parameters and feature subset \cite{Deb2002,Deb2001}. 

The fitness functions (Class $+1$ and Class $-1$ accuracy) are calculated for each solution (i.e., CKNN parameters and feature subset) by considering the average performance on cross-validation samples obtained using the leave-one-out method. This method has been shown to give good estimates of generalization ability \cite{vapnik98}, and would therefore help us in arriving at the best possible model from a deployment standpoint.

Since NSGA-II is a multi-objective optimization method, its purpose is to generate a \textit{Pareto frontier} of solutions (CKNN models), namely those which represent the best possible trade-off between the various objectives (Class $+1$ and Class $-1$ accuracy). Table \ref{tab:cknnres} gives a summary of the results.

\begin{table}[htbp]
  \caption{Citation Nearest Neighbour algorithm results arrived at using the NSGA-II optimization method}
    \begin{tabular}{|r|r|r|}
    \hline
    Class 0 accuracy & Class 1 accuracy & \# Models \\
    \hline \hline
    100\% & 91.49\% & 12 \\ \hline
    95.56\% & 95.74\% & 42 \\ \hline
    93.33\% & 97.87\% & 16 \\ \hline
    84.44\% & 100\% & 30 \\ \hline
    \end{tabular}%
  \label{tab:cknnres}%
\end{table}%

\subsection{Stacked ensemble}
\label{sec:stackedens}

We find that the results of the CNN algorithm, tuned as described in Section \ref{sec:exp-cnn} above, do not yet approach the performance level desired by us. We therefore consider using an ensemble approach, whereby we combine the predictions of the various CKNN models arrived at in the final generation of the NSGA-II run. Since these models approximate the Pareto frontier, it is possible that their combination would allow us to come up with a hybrid model that does even better on both objectives. Also, we wish to keep unrestricted, the method of combination of the CKNN predictions; therefore, we use the stacked generalization approach proposed in \cite{Wolpert1992}.

We therefore model the second level learning problem as one of mapping the predictions from the last generation of CKNN models to the desired sequence labels. We choose a Support Vector Machine classifier \cite{Cristianini2000,Hsu2000} with a Radial Basis Function kernel in order to combine the predictions. In order to optimize the $\gamma$ and $C$ parameters of the SVM model, as well as pick the optimal subset of CKNN models whose predictions are to be combined, we again use the NSGA-II algorithm as described in Section  \ref{sec:exp-cnn}.

\subsection{Experimental results}
\label{sec:expresult}

Since NSGA-II generates a Pareto frontier of solutions, a sample of three solutions of the stacked ensemble model is given in Table \ref{tab:stackensres}. These results suggest that the stacking layer improves the trade-off between accuracy on the two classes.

\begin{table}[htbp]
  \caption{Selection of solutions arrived at using the stacked ensemble}
    \begin{tabular}{|r|r|r|}
    \hline
    Class 0 accuracy & Class 1 accuracy & \# Models \\
    \hline \hline
    100\% & 93.61\% & 76 \\ \hline
    97.78\% & 100\% & 24 \\ \hline
    \end{tabular}%
  \label{tab:stackensres}%
\end{table}%

\section{Suggestions for further work}
\label{sec:concl}

The approach described here involves a number of components; therefore, further analysis needs to be done in order to better understand its applicability and correspondence to domain knowledge. From an algorithmic standpoint, one obvious area of further work for the paper is to test it on a diverse set of problems and benchmark it against methods such as those proposed in \cite{Dietterich1997,Zhou2007}, as a way of validating the effectiveness of the proposed method. 

Furthermore, we have noticed that, for problems with large feature sizes, the computational effort required to arrive at a solution can be quite high. Parallelism, caching and other tactics need to be explored further in order to make this a viable solution.

\bibliography{mil}

\end{document}